\pdfoutput=1
\PassOptionsToPackage{table,dvipsnames}{xcolor}
\documentclass[11pt]{article}
\usepackage[square, comma, sort&compress, numbers]{natbib}
\usepackage{ACL2023}
\usepackage{graphicx}
\usepackage{times}
\usepackage{latexsym}
\usepackage[T1]{fontenc}

\usepackage[T1]{fontenc}
\usepackage[utf8]{inputenc}

\usepackage{microtype}

\usepackage{inconsolata}
\usepackage{booktabs}
\usepackage{algorithm,algorithmicx}
\usepackage[noend]{algpseudocode}
\usepackage{multirow}
\usepackage{color}
\usepackage[table]{xcolor}
\usepackage{graphicx}
\usepackage{amsmath}
\title{Pushing The Limit of LLM Capacity for Text Classification}

\author{Yazhou Zhang$^{1,2}$, \ Mengyao Wang$^{2}$, Chenyu Ren$^1$\Thanks{Corresponding authors}, \ Qiuchi Li$^3$, \ Prayag Tiwari$^4$, \ Benyou Wang$^5$, \ Jing Qin$^{1*}$\\
    $^1$The Hong Kong Polytechnic University \\
    $^2$Zhengzhou University of Light Industry\\ 
    $^3$Copenhagen University \\
    $^4$Halmstad University\\
    $^5$The Chinese University of Hong Kong, Shenzhen \\
	} 
\begin{document}
\maketitle
\begin{abstract}
%Text classification has always been an enduring task in natural language processing (NLP). 
The value of text classification's future research has encountered challenges and uncertainties, due to the extraordinary efficacy demonstrated by large language models (LLMs) across numerous downstream NLP tasks.
In this era of open-ended language modeling, where task boundaries are gradually fading, an urgent question emerges: \textit{have we made significant progress in text classification with the full benefit of LLMs?}
To answer this question, 
we propose RGPT, an adaptive boosting framework tailored to produce a specialized text classification LLM by recurrently ensembling a pool of strong base learners. 
The base learners are constructed by adaptively adjusting the distribution of training samples and iteratively fine-tuning LLMs with them.
Such base learners are then ensembled to be a specialized text classification LLM, by recurrently incorporating the historical predictions from the previous learners.
Through a comprehensive empirical comparison, we show that RGPT significantly outperforms 8 SOTA PLMs and 7 SOTA LLMs on four benchmarks by 1.36\% on average. Further evaluation experiments reveal a clear superiority of RGPT over average human classification performance\footnote{Our codes are available at https://github.com/annoymity2024/RGPT\_2024}.
%Despite that general large language models (LLMs) have shown extraordinary efficacy across numerous downstream natural language processing (NLP) tasks, a significant concern revolves around their weak adaptation into specific tasks. General LLMs can be readily accessed from their open-source repositories, where strong specialized LLMs are difficult to train due to the need of large-scale task-specific knowledge and sophisticated reasoning abilities. To alleviate this gap, we propose AdaGPT, an adaptive boosting framework designed to produce a strong specialized LLM by cascading a collection of general LLMs. It mainly consists of two modules: weak adaptors and a combiner. The weak adaptors involve iteratively fine-tuning general LLMs with limited samples from specific tasks and thus adaptively adjusting the distribution of samples according the error rate of the current adaptor. Additionally, the combiner attentively integrates the above adaptors into series to be a strong specialized LLM. We offer a comprehensive evaluation of our proposed model on four benchmarking classification datasets and the resuls show that AdaGPT significantly outperforms 15 stateof-the-art baselines and 3 individual LLMs across various metrics. The results also show that AdaGPT benefits more to LLaMA 2 than another two LLMs.
\end{abstract}

\section{Introduction}
Text classification aims to assign pre-defined categories to a given informative text, including sentiment analysis, topic labeling, news classification, etc. It has always been an active task across the eras of knowledge engineering and feature engineering~\cite{cunha2023effective, minaee2021deep}. 
Recently, remarkable advances in LLMs, e.g., ChatGPT\footnote{https://chat.openai.com/}, GPT-4~\cite{openai2023gpt4}, ChatGLM 2~\cite{zeng2023glm-130b}, LLaMA 2~\cite{touvron2023llama}, etc., have 
demonstrated their outstanding performance across  downstream NLP tasks. Through instruction fine-tuning and in-context learning, LLMs have possessed marvelous language understanding, generation and reasoning abilities.

%Different from previous machine/deep learning approaches, text classification in LLM research is often treated as a simple label generation task, which has attracted the attention of only a very small minority.  

%  \begin{figure}[t]
%     \centering
%     \includegraphics[width=3.2in]{./images/example.png}
%   \caption{The comparison between the discriminative paradigm and generative paradigm.}
%    \label{fig:example}
% \end{figure}

Sustained efforts and investments from both academia and industry have been primarily dedicated to two directions: (1) general LLMs capable of providing encyclopaedic domain knowledge and performing well across a range of tasks, such as  Mistral~\cite{jiang2023mistral}, LLaMA series, etc.; (2) specialized LLMs tailored for vertical domains such as healthcare~\cite{chen2023huatuogpt,singhal2023large}, law~\cite{cui2023chatlaw}, finance~\cite{wu2023bloomberggpt}, education~\cite{milano2023large}, etc., via task-specific architectures and knowledge. Additionally, arming LLMs with strategies such as mixture-of-experts (MoE)~\cite{shen2023mixtureofexperts}, tool learning~\cite{qin2023tool} or modularization~\cite{ye2023mplugowl} have also garnered considerable attention. Strong LLMs intertwined with sophisticated optimization approaches are propelling LLM research to new heights.

%In the era of LLMs, text classification is often treated as a label generation task, where the model is used to generate the correct label-associated tokens. 
Despite the spotlight shining brighter on complicated tasks and exquisite domains, text classification languishes in the shadows with limited attention. Hence, an urgent research question emerges: 
%, as shown in Fig.~\ref{fig:example}.  As the study of LLMs increasingly focuses on complicated tasks and domains, there is limited attention given to text classification. Hence, an urgent research question emerges: 

%\textit{is text classification still an challenging task under the full encroachment of LLMs?} %it is important to investigate whether specialized text classification models can offer additional value beyond the existing LLM approaches

\textbf{RQ:} \textit{have we made significant progress in text classification with the full benefit of LLMs?} 

To answer this question, it is important to investigate whether specialized text classification LLM can create substantial value over the existing approaches.
We thus present \textbf{RGPT}, an adaptive boosting framework designed to investigate the limit of LLMs' classification ability. 
The main distinction from the recent text classification approaches, e.g., CARP~\cite{pavlopoulos2023detecting}, QLFR~\cite{wu2024quartet} and PromptBoosting~\cite{hou2023promptboosting} is that RGPT does not directly optimize the prompt space but instead builds a specialized LLM by adjusting sample distribution and recurrently ensembling strong base learners, thus demonstrating less sensitivity to prompts and stronger stability across various tasks (see Sec.~\ref{sec:setups} and \ref{Main Results}).

In particular, the base learners are constructed by iteratively fine-tuning LLMs with training samples. The distribution of training samples will be adaptively adjusted based on the error rates of the base learners. The misclassified samples will be given more weight, where the weights of correctly classified samples will be decreased.
Such base learners are then ensembled to be a specialized LLM, by taking the prediction and error rate of the previous learner as the contexts to prompt the current learner. 
This chain-like nature ensures that subsequent learners can improve and complement upon the existing knowledge.

We offer a comprehensive evaluation of the proposed RGPT model across four benchmark datasets and compare the results against 8 SOTA PLMs (e.g., DeBERTa, ERNIE, T5, etc.) and 7 SOTA LLMs (e.g., ChatGLM 2, LLaMA 2, GPT-4, etc.).
%, including RoBERTa, XLNet, RoBERTa-GCN, DeBERTa, ERNIE, T5, E2SC-IS, ContGCN and 7 SOTA LLMs, i.e., BBTv2, PromptBoosting, CARP, ChatGLM, LLaMA 2, QLFR and GPT-4.
The experimental results show the effectiveness of RGPT with the margin of 0.88\%, 1.21\%, 1.47\% and 1.88\% for four datasets. The study reveals that RGPT with only 7 iterations achieves the state-of-the-art results with performance continuing to grow as the number of iterations increases. Further human evaluation experiments demonstrate a clear surpassing of RGPT over average human classification.
A series of sub-experiments also prove that RGPT can universally boost varies base model structures.   
Hence, our study comes to a clear conclusion: our approach has pushed the limit of LLM capacity for text classification.
The main contributions are concluded as follows:
\begin{itemize}
\item We make the first attempt to explore the ongoing research value of text classification in the era of LLMs.

\item We propose RGPT, an adaptive boosting framework to push the limit of  LLMs’ classification ability.

\item Comprehensive experiments over four datasets demonstrate the effectiveness of RGPT in zero-shot text classification.
\end{itemize}

 \begin{figure*}[t]
    \centering
    \includegraphics[width=6in]{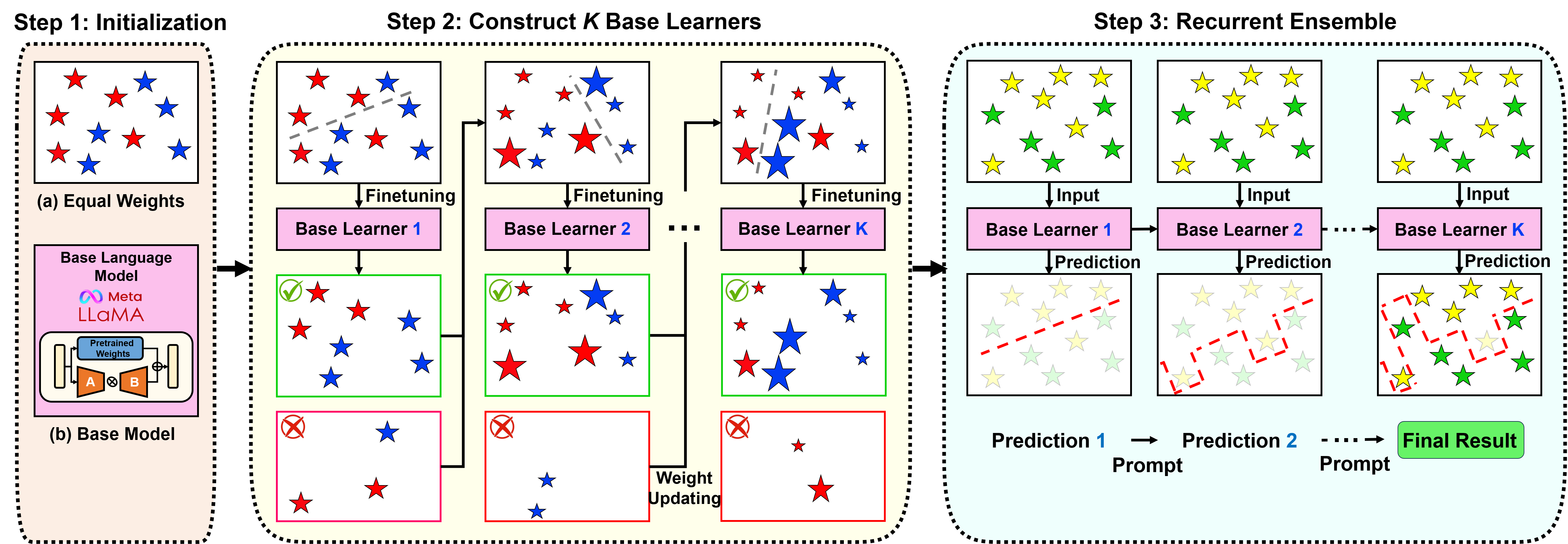}
  \caption{Overview of RGPT.}
   \label{fig:model}
\end{figure*}

\section{Preliminaries}
\subsection{Problem Definition}
Text classification is transformed as a conditional generative task, where the ouput $\mathcal{Y}$ will be the labels. Given a set of input documents $\mathcal{X} = \{x_1, x_2, \ldots, x_{N}\}$ where each document $x_i$ is augmented with a designed prompt ${Prompt}_i \in \mathcal{P}$ that provides contextual guidance, i.e., ${Prompt}_i =  {INS}_i \oplus x_i  $, where ${INS}_i$ represents the task instruction, $\mathcal{P}$ represents the prompt set. 
Our task is to learn a text classification LLM $\mathcal{M}(\theta)$ which maps an input document to its target label: $\mathcal{M}(\mathcal{X}, \mathcal{P}, \theta) \rightarrow \mathcal{Y}$, where $\mathcal{Y} = \{y_1, y_2, \ldots, y_{N}\}$ denotes the label sequence generated by the LLM $\mathcal{M}(\theta)$ based on its comprehension of the documents and the provided prompts and $y_i\in R^c$, where $c$ is the class of $y_i$. We formulate the classification problem as:
\begin{equation} 
\begin{aligned}
\resizebox{0.87\hsize}{!}{$
\mathcal{M}{\left ( \theta  \right ) } =\mathrm{arg}~\underset{c}{max}\prod_{i}Prob\left ( y_i=c|x_i, {Prompt}_i, \theta \right )  
$}
\end{aligned}
\end{equation}

\subsection{Algorithm Overview}
The recent LLM based approaches focus on elaborating prompts to improve classification performance. However, the performance gains from prompt engineering are limited, and the potential of classification performance for LLMs has not been fully investigated.

In contrast, RGPT is able to quickly generate a large pool of strong base learners through adjusting the distribution of training samples and fine-tuning LLMs, and proposes a recurrent ensembling approach to harnesses their complementarity, leading to improved effectiveness and generalization (see Sec. \ref{Main Results}). As shown in Fig.~\ref{fig:model}, RGPT consists of the following key steps.

\textbf{Step 1:} Initialization. Assign each training sample the same weight: $\frac{1}{N}$, and select a general LLM as initial base learner $\mathcal{LM}_0$\footnote{It has been proven that boosting can also effectively combine strong base learners~\cite{wyner2017explaining}.}.

\textbf{Step 2:} Constructing $K$ base learners $\mathcal{LM}_K$. The $k^{th}$ base learner, $\mathcal{LM}_k$, is optimized under its respective loss function, which is essentially a weighted loss over training samples with larger weights on those that are misclassified by the previous learner $\mathcal{LM}_{k-1}$.  

\textbf{Step 3:} Integrating $K$ base learners using a recurrent ensembling approach. More details will be provided in Sec.~\ref{sec:model} and Algorithm~\ref{alg:rgpt} in App.B.

\section{The Proposed Framework: RGPT}\label{sec:model}

\subsection{Initialization and Base Learner Selection}
To lay the groundwork for subsequent base learner construction and ensembling, we commence with initialization. 
Let $\mathcal{D}^{\left ( 0 \right ) } $ be the initial training set including $N$ samples. Each sample $(x_i^{\left ( 0 \right ) }, y_i^{\left ( 0 \right ) }) \in \mathcal{D}^{\left ( 0 \right ) }$, where $x_i^{\left ( 0 \right ) } \in \mathcal{X} $ is an input document and $y_i^{\left ( 0 \right ) } \in \mathcal{Y}$ its corresponding label. 

\textbf{(1) Weight initialization.} Suppose $\mathcal{W}^{\left ( 0 \right ) }=\left \{ w_1^{\left ( 0 \right ) }, w_2^{\left ( 0 \right ) },...,w_{N}^{\left ( 0 \right ) } \right \} $, where $\mathcal{W}^{\left ( 0 \right ) }$ represents the weight distribution of the initial  training samples. Each sample will be initialized as the same weight, i.e., $w_i^{\left ( 0 \right ) }= \frac{1}{N} $, where $\mathcal{W}^{\left ( 0 \right ) }\sim U\left (  \frac{1}{N}, \frac{1}{N},..., \frac{1}{N} \right )  $.
These weights will later be updated based on the error rate of the base learner.

\textbf{(2) Initial base learner selection.} In boosting, base learner can not only be a simple model (e.g., decision tree), but also be a strong learner that has yet considerable room to achieve optimal performance, such as DCNN~\cite{moghimi2016boosted}. %DBN~\cite{liu2014facial}, DCNN~\cite{moghimi2016boosted}, etc. 

%Due to the open-source nature of numerous LLMs, acquiring a well-pretrained LLM has become highly accessible, making it a suitable selection as an initial base learner.

We prove that our model works almost equally well on different base learners such as PLMs (i.e., RoBERTa) and LLMs (i.e., Alpaca\footnote{https://crfm.stanford.edu/2023/03/13/alpaca.html.},  LLaMA 2, ChatGLM 2). LLaMA 2 is selected as an initial base learner $\mathcal{LM}_0$, in view that it empirically yields the best result (see Sec.~\ref{sec:llms}).

% LLaMA 2-7B is an open source foundation language model, which is small and easy to reproduce. Additionally, it has been demonstrated inadequacy for addressing specialized tasks. Hence, we select it as the initial base learner, i.e., $\mathcal{LM}_0=LLaMA~2$. Other LLMs, i.e., Alpaca\footnote{https://crfm.stanford.edu/2023/03/13/alpaca.html.} and ChatGLM 2~\cite{du2022glm} are also attempted and compared in Sec.~\ref{sec:llms}. 

\subsection{Constructing Base Learners}
The construction of $K$ base learners involves (1) prompt construction; (2) fine-tuning LLMs with training samples; and (3) iteratively updating the weight distribution of training samples.

We follow the zero-shot prompting paradigm for text classification tasks. At each iteration $k$, the zero-shot prompt template ${Prompt}_i$ consists of two components: task instruction ${INS}_i$ and input document $x_i^{\left ( k \right ) }$. Task instruction ${INS}_i$ provides specifications for a text classification target and states the output constraint, e.g., ``\textit{Classify the SENTIMENT of the INPUT, and assign an accuracy label from [`Positive', `Negative'].} ''
%given an input document, $x_i^{\left ( k \right ) } \in \mathcal{X} $, the task is to generate a pre-defined label $y_i^{\left ( k \right ) }$ conditioning on the prompt ${Prompt}_i$ using a LLM. 
 
The $k^{th}$ base learner $\mathcal{LM}_k$ involves fine-tuning a general LLM using the training samples with the weight distribution, $\mathcal{W}^{\left ( k \right ) }=\left \{ w_1^{\left ( k \right ) }, w_2^{\left ( k \right ) },...,w_{N}^{\left ( k \right ) } \right \} $, effectively adjusting the model's focus on challenging samples. The objective is achieved by minimizing the weighted loss function:
\begin{gather}
\resizebox{0.87\hsize}{!}{$
\mathcal{LM}_k = \mathrm{arg}~\underset{\theta^{\left ( k \right ) }}{min} \sum_{\mathcal{D}^{\left ( k \right ) }} w_i^{\left ( k \right ) } \cdot \mathcal{L}(y_i^{\left ( k \right ) }, f(x_i^{\left ( k \right ) }; \theta^{\left ( k \right ) }))
$}
\end{gather}
where $\theta^{\left ( k \right ) }$ represents the parameters, $\mathcal{L}$ is the loss function, $f\left ( \cdot  \right ) $ is a general LLM (e.g., LLaMA 2).

Then, we compute its error rate $\epsilon^{\left ( k \right ) } $ and weight coefficient $\alpha ^{\left ( k \right ) }$, and thus update the distribution of training samples to guide the next iteration's focus towards misclassified samples:
\begin{equation} 
\resizebox{0.89\hsize}{!}{$\begin{aligned}
\epsilon^{\left ( k \right ) }&=Pr_{i\sim \mathcal{D}^{\left ( k \right ) }}\left [ \mathcal{LM}_k\left ( x_i^{\left ( k \right ) }  \right )\ne y_i^{\left ( k \right ) }  \right ]\\
 \alpha^{\left ( k \right ) }&=log\frac{1-\epsilon^{\left ( k \right ) }}{\epsilon^{\left ( k \right ) }} +log\left ( c-1 \right ) \\
 \mathcal{W}^{\left ( k+1 \right ) }&=\frac{\mathcal{W}^{\left ( k \right ) }}{Z_k}\times \begin{Bmatrix}
 e^{- \alpha^{\left ( k \right ) }} & if~\mathcal{LM}_k\left ( x_i^{\left ( k \right ) }  \right )=  y_i^{\left ( k \right ) } \\
 e^{ \alpha^{\left ( k \right ) }} & if~\mathcal{LM}_k\left ( x_i^{\left ( k \right ) }  \right )\ne   y_i^{\left ( k \right ) } 
\end{Bmatrix} 
\end{aligned}$}\label{eq:eq3}
\end{equation} 
where $c$ denotes the number of class, $Z_k$ represents the normalizing factor. Eq.~\ref{eq:eq3} will assign higher weights to samples with higher errors, and ensure that subsequent learners address the weaknesses of the current learner. 
After $K$ iterations, we construct $K$ complementary and strong base learners $\left \{ \mathcal{LM}_1, \mathcal{LM}_2,..., \mathcal{LM}_k \right \} $ (More explanations are provided in App.~\ref{sec:appa}).
%follow the SAMME algorithm~\cite{hastie2009multi},

 \begin{figure}[t]
    \centering
    \includegraphics[width=3.1in]{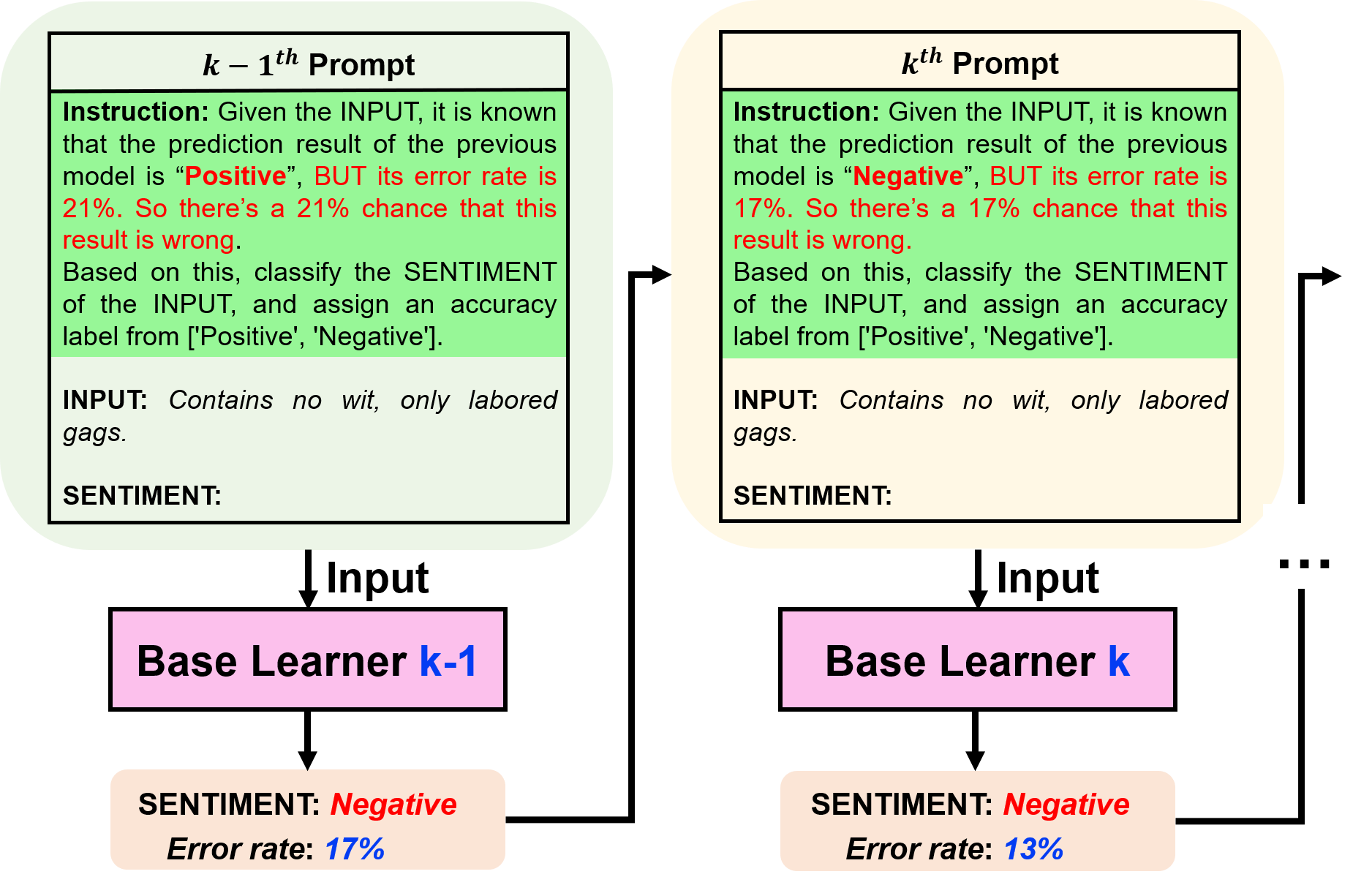}
  \caption{Recurrent ensembling $K$ base learners.}
   \label{fig:prompt}
\end{figure}

\subsection{Recurrently Ensembling the Base Learners}
%Instead of adopting linear combination or majority voting to directly ensemble $K$ base learners, 
We propose a recurrent ensembling approach, which selectively leverages the historical outputs generated by the previous learners. More specifically, the prediction result $\hat{y}^{k-1}_i$ of the previous learner $\mathcal{LM}_{k-1}$ along with its error rate $\epsilon^{\left ( k-1 \right ) }$ will be incorporated into the input prompt for the current learner $\mathcal{LM}_k$, which can be written as:
\begin{equation} 
\begin{aligned}
{Prompt}_i = {INS}_i \oplus {x}^{k}_i \oplus \{\hat{y}^{k-1}_i, \epsilon^{\left ( k-1 \right ) }\} 
\end{aligned}\label{eq:eq4}
\end{equation} 
where $\hat{y}^{k-1}_i$ is considered the supplementary knowledge for $\mathcal{LM}_k$. The error rate $\epsilon^{\left ( k-1 \right ) }$ acts as a trustworthiness metric, determining whether to rely on and adopt the prediction result of $\mathcal{LM}_{k-1}$, as shown in Fig.~\ref{fig:prompt}. 

This chain-like nature ensures that each subsequent learner can improve and complement upon the existing knowledge and producing a knowledge accumulation effect.
Finally, a strong, specialized LLM $\mathcal{M}(\theta)$ is constructed.
%. Further details are shown in Algorithm~\ref{alg:rgpt} in App.B.

\section{Experiments}
\subsection{Experiment Setups}\label{sec:setups}
\textbf{Datasets.} Four benchmarking datasets are selected as the experimental beds, $viz.$ SST-2~\cite{socher2013recursive}, MR~\cite{pang-etal-2002-thumbs}, AG News~\cite{zhang2015character}, Ohsumed\footnote{http://davis.wpi.edu/xmdv/datasets/ohsumed.html}. 
%from three classification tasks (i.e., sentiment classification, news classification and topic labeling)  
The statistics for each dataset are shown in Table~\ref{tab:statistics}.
%如果扩展期刊，这里应该对each一个数据集有1小段的介绍。

\begin{table}[t]
\centering
\footnotesize
\scalebox{0.9}{
\begin{tabular}{cccccc}
\toprule
\textbf{Dataset}  & \textbf{Task} & \textbf{Class} & \textbf{Avg. Length} & \textbf{\#Train}                                                             & \textbf{\#Test} \\
\toprule

SST-2    & Sentiment     & 2              & 17              & 6,920  & 1821                \\
MR    & Sentiment     & 2              & 20              & 8,662  & 2,000                \\
AG News & News          & 4              & 47            & 120,000 & 7,600               \\
Ohsumed  & Topic         & 23             & 136             & 3,357 & 4,043          \\
%SQuAD2.0 & Q\&A          & -              & 5000            & 130,319                                                                    & 11,873       \\
       \toprule  
\end{tabular}
}
\caption{Dataset statistics.} 
\label{tab:statistics}
\end{table}

\begin{table*}[t]
\centering
\small
\scalebox{0.94}{
\begin{tabular}{lcccccccc>{\columncolor{pink!20}}c>{\columncolor{yellow!10}}c}
\toprule
\multirow{2}{*}{\textbf{Method}} & \multicolumn{2}{c}{\textbf{SST-2}}                    & \multicolumn{2}{c}{\textbf{AG}}                       & \multicolumn{2}{c}{\textbf{Ohsumed}}                  & \multicolumn{2}{c}{\textbf{MR}}    & \multicolumn{2}{c}{\textbf{Avg. of Acc.}}                   \\ \cline{2-11}
                                 & Acc.                      & Ma-F1                     & Acc.                      & Ma-F1                     & Acc.                      & Ma-F1                     & Acc.                      & Ma-F1       & No Ohsumed       & All        \\ \midrule[1pt]

RoBERTa                          & 96.40                     & 96.23                     & 94.69                     & 94.35                     & 72.80                      &72.57                    & 89.42                     & -        & 93.50     & 88.32   \\

XLNet                            & 96.80                      & 96.67                      & 95.51                      & 95.18                      & 70.70                      & 70.41                   & 87.20                     & -          & 93.17       & 87.55        \\

RoBERTa-GCN                                & 95.80                      &-                      & 95.68  & -                      & 72.94                      & -                     &      89.70               & -           & 93.73    & 87.53            \\

DeBERTa                          &94.75                    & 94.15                      & 95.32                      & -                     & \textcolor{blue}{\underline{\textbf{75.94}}}                      & -                    & 90.21                      & 90.70                & 93.43         & \textcolor{blue}{\underline{\textbf{89.01}}}                           \\

ERNIE                           & \textcolor{blue}{\underline{\textbf{97.80}}}                      & -                     &-                     & - & 73.33                      &-                      & 89.53                    & -                      & -                      & -                      \\

T5-11B                           &97.50                     & 97.18                    & 92.21  & -    &51.72                      & 44.10                      &91.15                      & -            & 93.62             & 83.15 \\

\midrule
E2SC-IS                          & -                      & 93.10                     & -                     & 86.30 &-                     & 76.10                      & -                     & 88.60        & 89.33     & 86.02                             \\

%ATMIX                            & 92.67                      & -                      & -                     & -                      &-                     & -                      & -                     & -                    \\

ContGCN     & -                      & -                      & -                     & -                      &73.40 & -                      &91.30                     & -                 & -                      & -        \\

\midrule
BBTv2                            & 90.33                      & -                      & 85.28                      & -                      & -                      & -                     & 83.70                     & -                       & 86.44                     & -      \\

PromptBoosting                   & 87.60                      & -                     & 85.20   & -  & -   & -                      & 84.70                     &-                & 85.83                    &-       \\

CARP                             & 97.39                      & 97.14                      & \textcolor{blue}{\underline{\textbf{96.40}}}                      & -                    &  -                       &  -                       & \textcolor{blue}{ \underline{\textbf{92.39}}}                       & -             &\textcolor{blue}{\underline{\textbf{95.39}}}             & -\\

\midrule
ChatGLM 2                           & 81.36                      & 80.11                      & 83.67  & 83.67 & 54.33                      &41.84                                          & 74.39                    & 74.27          & 79.57                      & 74.01     \\

%ChatGLM                           & 79.40                      & 78.23                      & 82.51  & 82.50 & 51.28                      &36.68                                          & 73.50                    & 73.31          & 78.46                      & 71.67     \\
 
LLaMA 2                         & 60.50                     & 61.08                      & 79.40                      & 80.67                      & 48.08                      & 40.21                     & 71.49                      & 71.03                   & 62.69                      & 64.89       \\

QLFR                          &-                      & -                    & 89.14                     & 89.28 & 61.10                      & 51.85                      & 81.70                      & 81.72                     &-                      & -      \\ 

GPT-4                          & 82.52                      & 81.17                      & 84.62                      & 84.50                      & 55.20                      & 51.26                      & 77.90                      & 77.63                        & 81.68                      & 75.06     \\

\midrule
\rowcolor{gray!20} RGPT                             & \textbf{{98.68}$_{\pm0.2}$}             &  \textbf{{98.67}}             &  \textbf{{97.61}$_{\pm0.3}$}             &  \textbf{{97.52}}             &  \textbf{{77.41}$_{\pm0.2}$}            &  \textbf{{73.68}}            &  \textbf{{94.27}$_{\pm0.5}$}      &           \textbf{{94.15}}     &  \textbf{{96.85}}      &           \textbf{{91.99}}  \\

\rowcolor{gray!20} Gain~$\bigtriangleup $                             & \multicolumn{1}{l}{0.88\%} & \multicolumn{1}{l}{1.49\%} & \multicolumn{1}{l}{1.21\%} & \multicolumn{1}{l}{2.34\%} & \multicolumn{1}{l}{1.47\%} & \multicolumn{1}{l}{0.76\%} & \multicolumn{1}{l}{1.88\%} & \multicolumn{1}{l}{3.45\%}    & 1.46\% & 2.98\%   \\

\bottomrule[1pt]
\end{tabular}
}
\caption{Performance on four datasets. Bold and \textcolor{blue}{blue} indicate the best and second-best results for each dataset.}
\label{tab:baseline}
\end{table*}

\textbf{Baselines.} A wide range of SOTA baselines are included for comparison. They are: (1) \underline{\textbf{RoBERTa}}~\cite{liu2019roberta}, (2) \underline{\textbf{XLNet}}~\cite{yang2019xlnet}, (3) \underline{\textbf{RoBERTa-GCN}}~\cite{lin-etal-2021-bertgcn}, (4) \underline{\textbf{DeBERTa}}~\cite{he2020deberta}, (5) \underline{\textbf{ERNIE}}~\cite{sun2021ernie} and (6)  \underline{\textbf{T5}}~\cite{raffel2020exploring} are six strong PLMs for text classification via masked language modeling and pretrained representations. (7) \underline{\textbf{E2SC-IS}}~\cite{cunha2023effective} selects the most representative documents for training classification model. (8) \underline{\textbf{ContGCN}}~\cite{Yao2018GraphCN} focuses on the misclassifed training samples as the target for explainable text classification. (9) \underline{\textbf{BBTv2}}~\cite{sun2022bbtv2}, (10) \underline{\textbf{PromptBoosting}}~\cite{hou2023promptboosting} and (11) \underline{\textbf{CARP}}~\cite{pavlopoulos2023detecting} are three SOTA prompt  based approaches that focus on how to find the best prompts given a specific classification task. (12) \underline{\textbf{ChatGLM 2}},  (13) \underline{\textbf{LLaMA 2}} and (14) \underline{\textbf{GPT-4}} are three SOTA LLMs that have broad domain knowledge and outstanding performance across various NLP tasks. (15) \underline{\textbf{QLFR}}~\cite{wu2024quartet} decomposes the text classification task into four distinct reasoning steps and presents a fine-tuned LLaMA 2-13B model.

\textbf{Implementation.} Training a base learner will cost about 1 hours on 8 $\times$ A100-SXM4-40GB GPUs. The micro batch size, batch size, the number of epoch and learning rate are set to be 8, 128, 10 and 3e-4 respectively. In the process of updating sample weights, we control the weights of samples by increasing or decreasing the number of samples. For a misclassified sample $x_i^k$, whose weight should increase to $w_i^{k+1}$ (see Eq.3), we proportionally augment its quantity. To improve generalization and avoid overfitting, we utilize ChatGPT to generate additional samples similar to $x_i^k$.
%for this purpose.
%We use low-rank adaptation (LoRA) to finetune base learner with only 2.1 million trainable parameters.

%\textbf{Evaluation metrics.} We adopt accuracy (Acc) and macro-F1 (Ma-F1) as evaluation metrics for a fair comparison.

\subsection{Main Results} \label{Main Results}
We report both \textbf{Accuracy} and \textbf{Macro-F1} results for RGPT and baselines in a zero-shot setting in Table~\ref{tab:baseline}. The mean and variance
over 5 runs are calculated. 
We observe that RGPT consistently achieves state-of-the-art performance on four datasets, i.e., 0.88\%$\uparrow$, 1.21\%$\uparrow$, 1.47\%$\uparrow$, 1.88\%$\uparrow$ respectively. It outperforms PLMs based, prompt based and standard fine-tuning approaches.
Despite that LLMs (i.e., ChatGLM 2, LLaMA 2, GPT-4) have shown extraordinary efficacy across general-domain tasks, their weak adaptation into text classification is also proved, in view of their worst classification performance. Among them, GPT-4 performs better than another two. By fine-tuning LLaMA 2-13B or optimizing in prompt space, QLFR, BBTv2, PromptBoosting and CARP gain significant improvements over general LLMs. QLFR, BBTv2 and PromptBoosting have been trading victories on different benchmarks, but they are inferior to other methods using PLMs, e.g., RoBERTa, DeBERTa, T5, etc. CARP achieves the best performance on AG News and MR datasets among all the baselines, and obtain comparable results against ERNIE on SST-2 dataset. This suggests that prompt learning indeed elicits LLMs to outperform traditional PLMs based approaches, but the design of prompts is critically important.

\begin{table}[t]
\centering
\footnotesize
\scalebox{0.89}{
\begin{tabular}{lcccc}
\toprule
\textbf{Method}  & \textbf{SST-2} & \textbf{AG News} & \textbf{Ohsumed}                                                             & \textbf{MR} \\
\toprule

\textit{w/o} Boosting    & 89.23     & 90.53              &67.73           & 88.08                \\
\textit{w/o} LLM    & 97.47     & 95.84              &74.70           & 93.28                  \\
\textit{w/o} Recurrent ensemble  &  98.18     & 96.90              &76.99           & 93.71               \\
\toprule 
RGPT  & 98.68     & 97.61              &77.41           & 94.27            \\

       \toprule  
\end{tabular}
}
\caption{Ablation study in a zero-shot setting.} 
\label{tab:ablation}
\end{table}

\subsection{Ablation Study}
Table~\ref{tab:ablation} shows the result of ablation studies on four datasets. For \textit{w/o Boosting}, we choose to directly fine-tune LLaMA 2-7B with initial training samples, removing the boosting strategy. For \textit{w/o LLM}, we substitute LLaMA 2-7B with small language model (namely RoBERTa) to be the backbone language model. For \textit{w/o Recurrent ensemble}, we perform a weighted combination of $K$ strong base learners according their coefficients $\alpha ^{\left ( k \right ) }$. From the experiment results above, we highlight the following conclusions: (a) boosting LLM making the greatest contribution in improving the classification performance; (b) LLMs demonstrating greater advancedness over PLMs for text classification; (c) the effectiveness of our proposed recurrent ensembling approach. In a summary, each module in our method contributes to the final performance.

 \begin{figure}[t]
    \centering
    \includegraphics[width=3.2in]{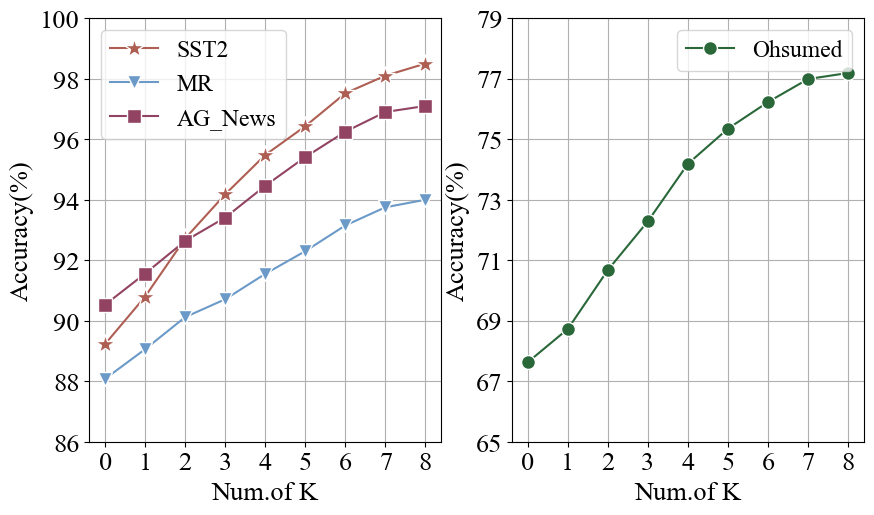}
  \caption{Performance of RGPT with increasing number of learners.}
   \label{fig:Keffect}
\end{figure}

\subsection{Effect of $K$}
In our main experiments, we adopt $K=7$ due to its significant SOTA performance. Intuitively, a large learners pool increases the diversity of base learners which could improve the performance. We empirically present the relationship between the number of learners and the model performance in Fig.~\ref{fig:Keffect}. As we have discussed in Table~\ref{tab:ablation}, an individual fine-tuned LLM performs very poorly (i.e., 83.89\% accuracy on average). However, by using our recurrent boosting framework, the performance can be boosted to 90.67\% when 6 base learners are provided, which slightly overcomes all the baselines. Further, when $K=7$, the performance can be boosted to 91.99\%, which significantly outperforms others with performance continuing to grow as the number of iterations increase (e.g., $K=8$). But the performance increase plateaus as the number of base learners rises from 7 to 8, suggesting that 7 base learners makes a good balance between performance and training cost. 
%However, the training cost will also increase where more base learners are included for fine-tuning. 

\begin{table}[t]
\centering
\footnotesize
\scalebox{0.9}{
\begin{tabular}{lcccc}
\toprule
\textbf{Method}  & \textbf{SST-2} & \textbf{AG News} & \textbf{Ohsumed}                                                             & \textbf{MR} \\
\toprule
RoBERTa    & 96.40     & 94.69              &72.80          & 89.42                \\
$\text{RGPT}_{+RoBERTa}$    & 97.47     & 95.84              &74.70           & 93.28                \\\toprule

Alpaca    & 57.81    & 71.23             &46.55          & 53.78                \\
$\text{RGPT}_{+Alpaca}$    & 97.81     & 96.45              &75.26           & 93.55                \\\toprule

ChatGLM~2    & 81.36     & 83.67              &54.33          & 74.39                \\
$\text{RGPT}_{+ChatGLM~2}$      & 98.10     & 96.77              &75.16           & 93.02                  \\
\toprule 

LLaMA~2    & 60.50    & 79.40              &48.08          & 71.49                \\
$\text{RGPT}_{+LLaMA~2}$  & 98.68     & 97.61              &77.41           & 94.27            \\

       \toprule  
\end{tabular}
}
\caption{The impact of different base learners.} 
\label{tab:varies}
\end{table}

\subsection{How RGPT Varies With Different Base Learners}\label{sec:llms}
We select LLaMA 2-7B to the initial base model by default. In order to evaluate the effect of different base learners, we have also tried another two SOTA LLMs and one strong PLM, i.e., Alpaca, ChatGLM 2 and RoBERTa, as shown in Table~\ref{tab:varies}. We notice that RGPT+RoBERTa performs the worst on four tasks, but still significantly outperforms the standard RoBERTa with the margin of 2.26\% on average. Additionally, RGPT+Alpaca obtains slightly improvements over RGPT+RoBERTa, but is inferior to ChatGLM 2 and LLaMA 2. The reason is that latter models have adopted more advanced architectures and training methodologies. In addition, three standard SOTA LLMs perform very poorly without boosting, which implies that general LLMs are still insufficient to directly cope with various text classification tasks. But their performance significantly improves using RGPT, with an increase of over 21.0\%$\uparrow$. Different base models can achieve comparable results using RGPT. We demonstrate that RGPT universally boosts varies base model structures.

\subsection{Zero-shot v/s Few-shot Prompting}\label{sec:fewshot}
We perform zero-shot and few-shot experiments to evaluate whether RGPT can perform better when a limited number of contextual examples are available. The results are shown in Table~\ref{sec:Few-shot}. We design four $k$-shot settings: zero-shot, one-shot, five-shot, ten-shot. For each setting, we randomly sample $k= \left \{ 0,1,5,10 \right \} $ examples from the training set. 

The impact of adding shots varies with the number of shots. The change from zero-shot to one-shot results in a slight improvement in classification performance. With the gradual increase in the number of shots, the performance drops down. This potentially arises from RGPT learning redundant information when handling too long contextual data. This implies that crudely increasing the number of extra shots does not necessarily result in a stable performance improvement.

\begin{table}[h]
\centering
\small
\scalebox{0.95}{
\begin{tabular}{c|cccc}
\toprule
\textbf{Prompt} & \textbf{SST-2} &  \textbf{AG News} & \textbf{Ohsumed}   & \textbf{MR}\\
 \midrule[1pt] 
%0-shot                  &98.20   &97.24   &76.92     & 93.95  \\ 
0-shot                  &98.68   &97.61   &77.41     & 94.27  \\ 

1-shot                  &98.97   &98.01   &77.83     & 94.65  \\ 
3-shot                 &98.31   &97.57   &77.32     & 94.11  \\ 
10-shot                &97.95   &96.60  &76.85     & 93.52  \\ \midrule[1pt] 
\end{tabular}}
\caption{Few shot performance testing.}\label{sec:Few-shot}
\end{table}

\subsection{Human v/s Machine} 
\begin{table}[t]
\small
\centering
\scalebox{0.95}{
\begin{tabular}{ccc}
\midrule[1pt] 
\textbf{Method} & \textbf{Accuracy} & \multicolumn{1}{c}{\textbf{Efficiency (minutes)}} \\  \midrule[1pt]
Human 1 &   89.21                    &       53.3   \\
         Human 2              &  90.05                    &                 56.9           \\
          Human 3               &  96.59                    &       80.6              \\
              Avg.              &  91.95                    &      63.6               \\ \midrule[1pt]
          RGPT              &  92.54                    &       10.9                 \\\midrule[1pt]     
\end{tabular}
}
\caption{ The human classification results against RGPT.}\label{tab:human}
\end{table}

We create a new test set including 200 samples  randomly sampled from three datasets, e.g., IMDB~\cite{maas2011learning}, R8\footnote{https://www.cs.umb.edu/~smimarog/textmining/datasets/} and DBPedia~\cite{auer2007dbpedia}, where their proportion is 4:3:3. Then, we recruit three volunteers\footnote{They all signed on the consent form before the study and were
paid an equal \$5.0/hour. Prior to annotation, they received professional guidance covering the criteria for labeling, positive and negative examples, etc.} to evaluate the sentiment, news and topic labels. We ask the first two annotators to proceed at their standard speeds, where the third annotator should annotate meticulously and conduct a double-check.
Their classification scores and time costs will be compared with RGPT in Table~\ref{tab:human}. 
It can be seen that RGPT consistently outperforms two humans in terms of accuracy and efficiency. Despite that RGPT underperforms the third annotator, its time cost is $\frac{1}{7} $ of that of the third annotator. It is foreseeable that with the continuous improvement of future LLMs, their classification capabilities will further enhance. RGPT also surpasses the average performance of three annotators, proving that we have made much progress in text classification over the existing approaches.

 \begin{figure}[t]
    \centering
    \includegraphics[width=3.0in]{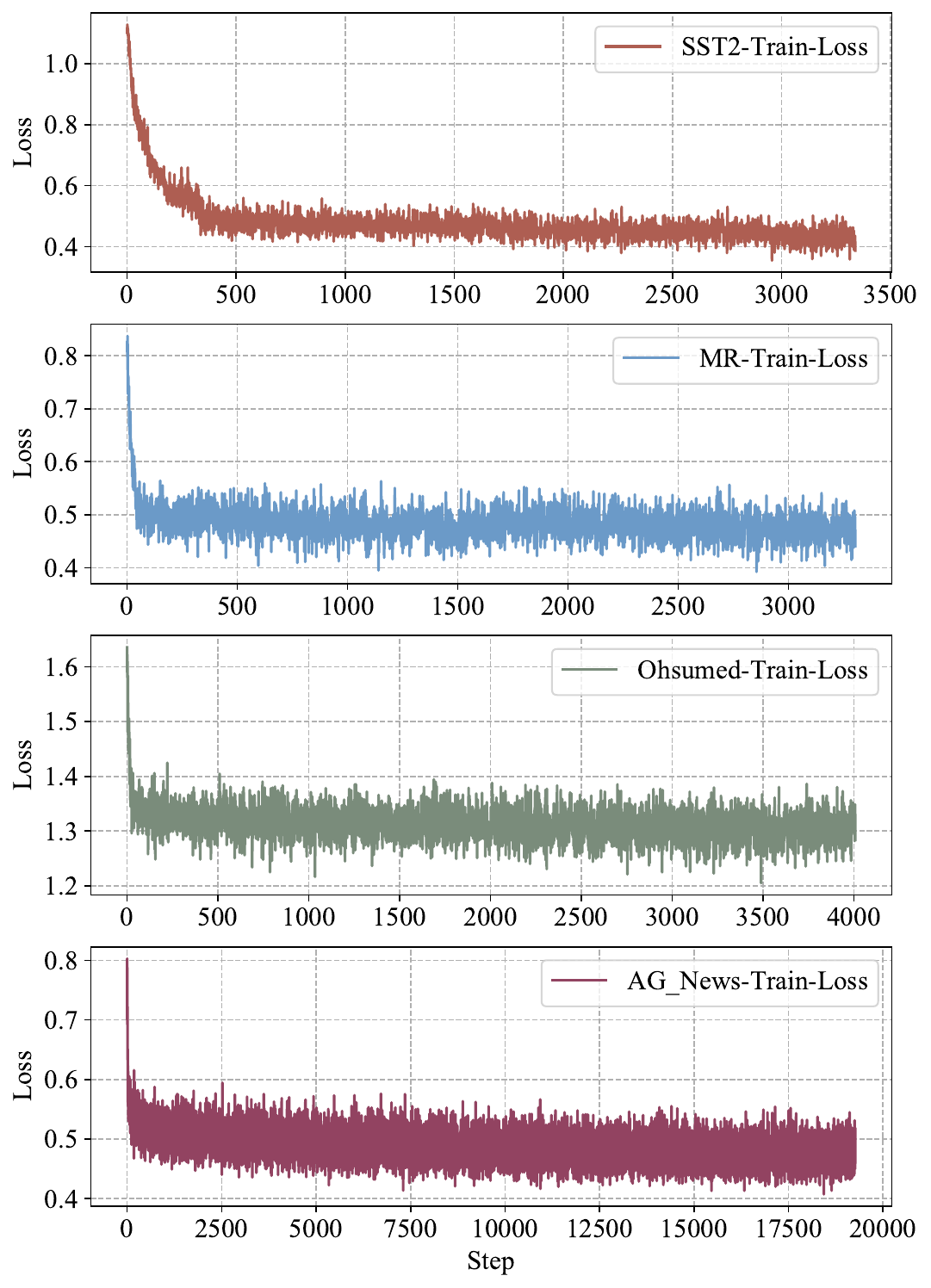}
  \caption{The training loss of RGPT.}
   \label{fig:Loss}
\end{figure}

\subsection{Overfitting Study}
To confirm that our model doesn't overfit when consistently adjusting the sample distribution, we adopt three strategies: (1) early stopping approach is used; (2) we present the learning curves to show how the training loss changes, as shown in Fig.~\ref{fig:Loss}. This visual representation helps us understand if the model's performance improves consistently on both new and seen data during training; (3) in addition, we do not directly increase (e.g., replicate) the number of those misclassified samples. Instead, we choose to increase similar style samples generated by ChatGPT. This strategy can improve the diversity of samples to avoid overfitting.
 
\begin{figure*}[t]
    \centering
    \includegraphics[width=5.9in]{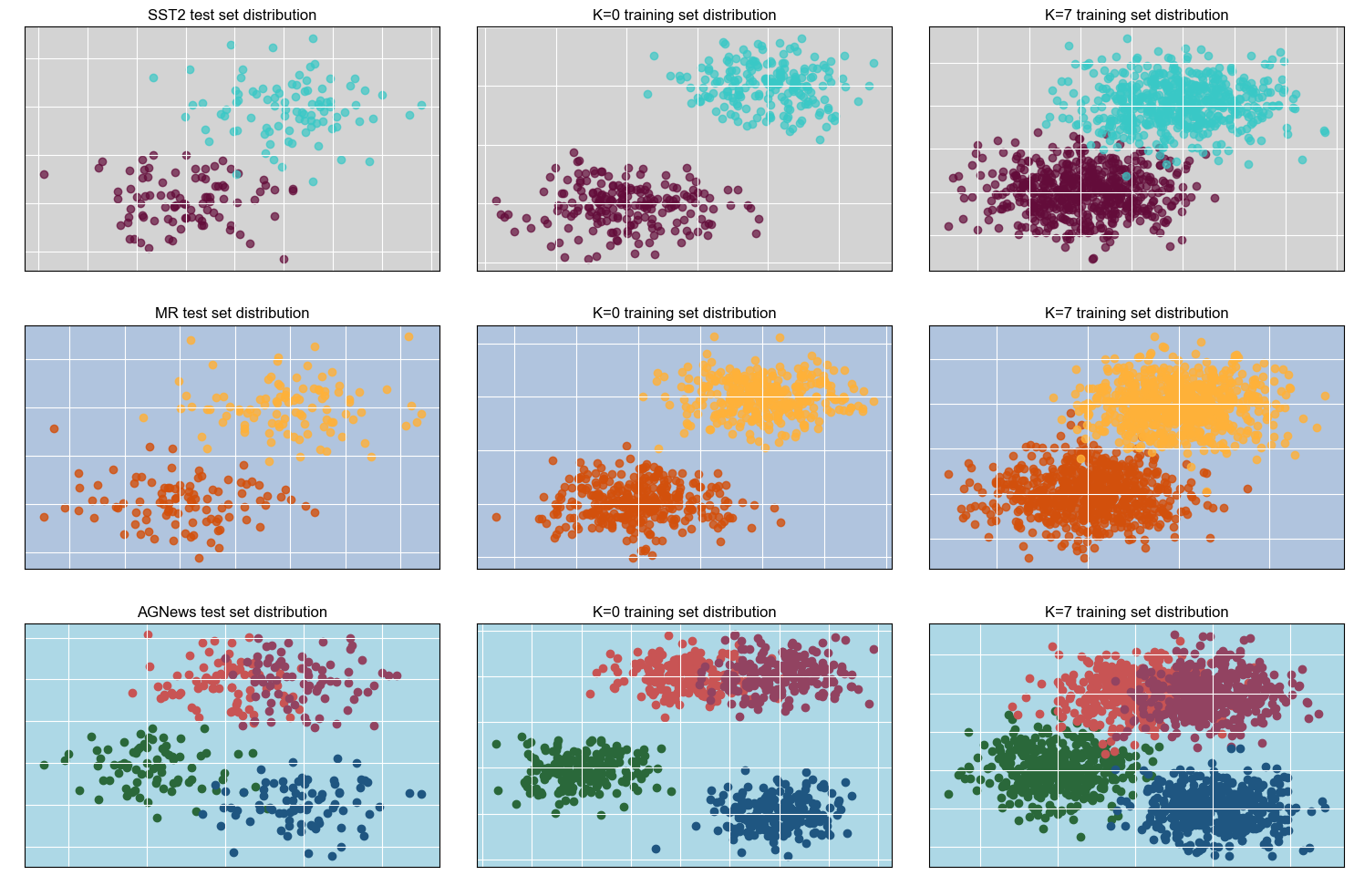}
  \caption{Distribution of training samples and initial test samples during K iterations.}
   \label{fig:Kiteration}
\end{figure*}

\subsection{Data Visualization}
We present a visual comparison chart between the distribution of testing set and the distribution of training set after $K=7$ iterations, as shown in Fig.~\ref{fig:Kiteration}.
We notice that the distribution of the training set at $K=0$ differs significantly from the test set distribution, while at $K=7$, the distribution of the training set becomes more aligned with the distribution of the testing set.
This indicates that our RGPT method effectively adjusts the distribution of the training set to be more similar to the true distribution, thereby enhancing the classification performance of the model. In addition, the distribution of the training set, evolving through iterative adjustments in boosting, exhibits concentration around previously misclassified samples, indicating the algorithm's focus on challenging cases.
The visualization provides a nuanced understanding of how the model adjusts the training data. This analysis aids in assessing the model's potential overfitting tendencies, and its ability to generalize effectively to new instances.

\subsection{Error Analysis}\label{sec:error}

\begin{figure}[htp]
    \centering
    \includegraphics[width=3.1in]{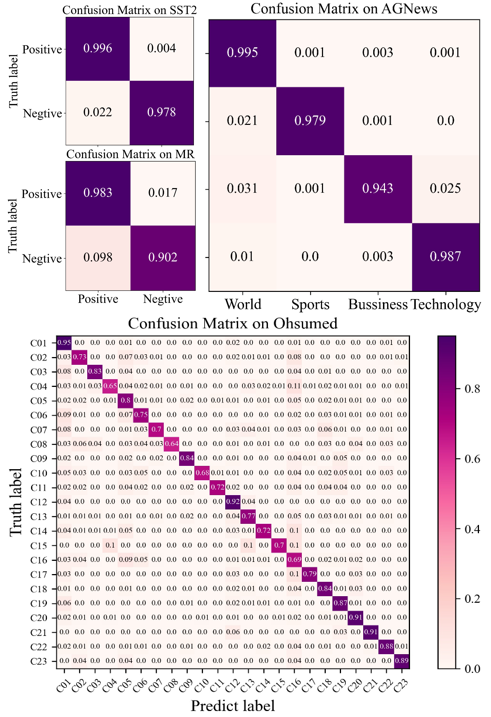}
  \caption{The normalized confusion matrices for RGPT across four datasets. The columns represent the truth label, where the rows represent the predicted labels.}
   \label{fig:confusionmatrix}
\end{figure}

The detailed error analysis is also conducted via the confusion matrices that are shown in Figure~\ref{fig:confusionmatrix}. Each cell $\left ( i,j \right ) $ represents the percentage of class $i$ is classified to be class $j$. Upon reviewing the classification results produced by RGPT on four datasets, we discover that imbalanced categories and the similarity across different categories are the key factors contributing to misclassification. 

By examining the diagonal elements of the matrices, RGPT demonstrates effective true-positive categorization for most fine-grained categories across four datasets. However, it exhibits a tendency to misclassify the ``negative'' utterances to be ``positive'', particularly on the SST-2 and MR datasets. In addition, RGPT tends to misclassify ``Bussiness'' to be ``World'' and ``Technology'' on AGNews dataset. RGPT has high error rate on Ohsumed dataset. There are two possible reasons: (1) the highly unbalanced samples leads to the model's misclassification, e.g., C18, C20, etc.; (2) the similarity across several categories, e.g., C4, C11, C12, C13, etc., may  pose a challenge for the model to accurately distinguish them.

\section{Related Work}\label{relatedwork}
%We provide a brief overview of the related research.

%\subsection{Large Language Models}
In recent years, significant advancements in NLP have been attributed to the emergence of LLMs. 
%These models have demonstrated remarkable capabilities such as in-context learning, few-shot prompting, instruction following, etc. 
OpenAI has achieved significant milestones with the creation of two groundbreaking models: ChatGPT and GPT-4. However, due to their proprietary nature, 
There has been numerous LLM variants featuring tens or even hundreds of billions of parameters~\cite{zhao2023survey}. We categorize these LLMs into two groups based on their specialization: general LLMs and specialized LLMs. General LLMs are designed for versatility across a wide spectrum of NLP tasks. Prominent examples of these models are GPT-4, ChatGLM, LLaMA 2, PanGu-$\Sigma$~\cite{ren2023pangu}, Falcon~\cite{penedo2023refinedweb}, etc. In contrast, specialized LLMs are fine-tuned for specific tasks via task-specific architectures and knowledge, allowing them to achieve higher  performance. An increasing number of studies are raging over medical, law, finance and education domains, e.g., HuaTuo~\cite{zhang2023huatuogpt}, FinGPT~\cite{yang2023fingpt}, ChatLaw~\cite{cui2023chatlaw}, etc. 

Different from the above-mentioned studies, we pioneer a specialized LLM by iteratively refining and integrating base LLMs, unlocking its untapped potential on text classification tasks.

\section{Conclusions}
In this work, we propose RGPT, an adaptive boosting framework tailored to produce a specialized text classification LLM. we efficiently train a pool of strong base learners by adjusting the distribution of training samples and iteratively fine-tuning LLMs with them. Such base learners are then recurrently ensembled to be a specialized LLM. We offer a comprehensive evaluation and our model achieves the state-of-the-art results. This proves that boosting LLMs will yield significant improvements over other PLM and prompt based approaches. Human evaluation experiments proves that RGPT can outperform average human performance. 
%We can draw a clear conclusion: our approach indeed pushes the limit of LLM capacity for text classification. 
%In the future, we plan to propose an unified model to advance all text classification tasks.
% is also left to future work. 
%investigate the effectiveness of RGPT across more text classification tasks. How to

\section{Limitations.} 
The proposed RGPT model also has several limitations: (1) \textbf{High computational cost.} The iterative nature of its boosting-based mechanism, which involves multiple rounds of fine-tuning LLMs, leads to a significant computational cost. (2)  \textbf{Limited testing sets.} RGPT has shown significant performance improvements across four benchmark datasets. However, the study does not thoroughly examine how well the model may work on a wider range of text classification tasks. (3) \textbf{Monotony of base learners.} Base learner should not only be homogeneous, but also can be heterogeneous. Limiting the RGPT framework's base learners solely to LLaMA 2 may hinder the method's innovation and its potential for improvement. Ensembling different LLMs may enhance the adaptability and versatility of the approach when facing new challenges.

\textbf{Potential Risks.} 
Even though RGPT addresses overfitting by increasing similar samples instead of the misclassified samples themselves, there remains a risk of overfitting during the repeated fine-tuning of large language models. This risk becomes more prominent in situations with a small training set.

%\section*{Acknowledgements}
%
%This document has been adapted
%by Steven Bethard, Ryan Cotterell and Rui Yan
%from the instructions for earlier ACL and NAACL proceedings, including those for 
%ACL 2019 by Douwe Kiela and Ivan Vuli\'{c},
%NAACL 2019 by Stephanie Lukin and Alla Roskovskaya, 
%ACL 2018 by Shay Cohen, Kevin Gimpel, and Wei Lu, 
%NAACL 2018 by Margaret Mitchell and Stephanie Lukin,
%Bib\TeX{} suggestions for (NA)ACL 2017/2018 from Jason Eisner,
%ACL 2017 by Dan Gildea and Min-Yen Kan, 
%NAACL 2017 by Margaret Mitchell, 
%ACL 2012 by Maggie Li and Michael White, 
%ACL 2010 by Jing-Shin Chang and Philipp Koehn, 
%ACL 2008 by Johanna D. Moore, Simone Teufel, James Allan, and Sadaoki Furui, 
%ACL 2005 by Hwee Tou Ng and Kemal Oflazer, 
%ACL 2002 by Eugene Charniak and Dekang Lin, 
%and earlier ACL and EACL formats written by several people, including
%John Chen, Henry S. Thompson and Donald Walker.
%Additional elements were taken from the formatting instructions of the \emph{International Joint Conference on Artificial Intelligence} and the \emph{Conference on Computer Vision and Pattern Recognition}.


\begin{thebibliography}{48}
\expandafter\ifx\csname natexlab\endcsname\relax\def\natexlab#1{#1}\fi

\bibitem[{Auer et~al.(2007)Auer, Bizer, Kobilarov, Lehmann, Cyganiak, and Ives}]{auer2007dbpedia}
S{\"o}ren Auer, Christian Bizer, Georgi Kobilarov, Jens Lehmann, Richard Cyganiak, and Zachary Ives. 2007.
\newblock Dbpedia: A nucleus for a web of open data.
\newblock In \emph{international semantic web conference}, pages 722--735. Springer.

\bibitem[{Chen et~al.(2023{\natexlab{a}})Chen, Wang, Gao, Jiang, Chen, Zhang, Song, Xie, Kong, Li et~al.}]{chen2023huatuogpt}
Junying Chen, Xidong Wang, Anningzhe Gao, Feng Jiang, Shunian Chen, Hongbo Zhang, Dingjie Song, Wenya Xie, Chuyi Kong, Jianquan Li, et~al. 2023{\natexlab{a}}.
\newblock Huatuogpt-ii, one-stage training for medical adaption of llms.
\newblock \emph{arXiv preprint arXiv:2311.09774}.

\bibitem[{Chen et~al.(2023{\natexlab{b}})Chen, Jiang, Chen, Wang, Yu, Chen, Zhang, Liang, Zhang, Zhang et~al.}]{chen2023phoenix}
Zhihong Chen, Feng Jiang, Junying Chen, Tiannan Wang, Fei Yu, Guiming Chen, Hongbo Zhang, Juhao Liang, Chen Zhang, Zhiyi Zhang, et~al. 2023{\natexlab{b}}.
\newblock Phoenix: Democratizing chatgpt across languages.
\newblock \emph{arXiv preprint arXiv:2304.10453}.

\bibitem[{Cui et~al.(2023)Cui, Li, Yan, Chen, and Yuan}]{cui2023chatlaw}
Jiaxi Cui, Zongjian Li, Yang Yan, Bohua Chen, and Li~Yuan. 2023.
\newblock Chatlaw: Open-source legal large language model with integrated external knowledge bases.
\newblock \emph{arXiv preprint arXiv:2306.16092}.

\bibitem[{Cunha et~al.(2023)Cunha, Fran{\c{c}}a, Fonseca, Rocha, and Gon{\c{c}}alves}]{cunha2023effective}
Washington Cunha, Celso Fran{\c{c}}a, Guilherme Fonseca, Leonardo Rocha, and Marcos~Andr{\'e} Gon{\c{c}}alves. 2023.
\newblock An effective, efficient, and scalable confidence-based instance selection framework for transformer-based text classification.
\newblock In \emph{Proceedings of the 46th International ACM SIGIR Conference on Research and Development in Information Retrieval}, pages 665--674.

\bibitem[{Du et~al.(2022)Du, Qian, Liu, Ding, Qiu, Yang, and Tang}]{du2022glm}
Zhengxiao Du, Yujie Qian, Xiao Liu, Ming Ding, Jiezhong Qiu, Zhilin Yang, and Jie Tang. 2022.
\newblock Glm: General language model pretraining with autoregressive blank infilling.
\newblock In \emph{Proceedings of the 60th Annual Meeting of the Association for Computational Linguistics (Volume 1: Long Papers)}, pages 320--335.

\bibitem[{Han et~al.(2022)Han, Zhao, Ding, Liu, and Sun}]{han2022ptr}
Xu~Han, Weilin Zhao, Ning Ding, Zhiyuan Liu, and Maosong Sun. 2022.
\newblock Ptr: Prompt tuning with rules for text classification.
\newblock \emph{AI Open}, 3:182--192.

\bibitem[{He et~al.(2020)He, Liu, Gao, and Chen}]{he2020deberta}
Pengcheng He, Xiaodong Liu, Jianfeng Gao, and Weizhu Chen. 2020.
\newblock Deberta: Decoding-enhanced bert with disentangled attention.
\newblock \emph{arXiv preprint arXiv:2006.03654}.

\bibitem[{Hou et~al.(2023)Hou, O'connor, Andreas, Chang, and Zhang}]{hou2023promptboosting}
Bairu Hou, Joe O'connor, Jacob Andreas, Shiyu Chang, and Yang Zhang. 2023.
\newblock Promptboosting: Black-box text classification with ten forward passes.
\newblock In \emph{International Conference on Machine Learning}, pages 13309--13324. PMLR.

\bibitem[{Jiang et~al.(2023)Jiang, Sablayrolles, Mensch, Bamford, Chaplot, de~las Casas, Bressand, Lengyel, Lample, Saulnier, Lavaud, Lachaux, Stock, Scao, Lavril, Wang, Lacroix, and Sayed}]{jiang2023mistral}
Albert~Q. Jiang, Alexandre Sablayrolles, Arthur Mensch, Chris Bamford, Devendra~Singh Chaplot, Diego de~las Casas, Florian Bressand, Gianna Lengyel, Guillaume Lample, Lucile Saulnier, Lélio~Renard Lavaud, Marie-Anne Lachaux, Pierre Stock, Teven~Le Scao, Thibaut Lavril, Thomas Wang, Timothée Lacroix, and William~El Sayed. 2023.
\newblock \href {http://arxiv.org/abs/2310.06825} {Mistral 7b}.

\bibitem[{Kenton and Toutanova(2019)}]{kenton2019bert}
Jacob Devlin Ming-Wei~Chang Kenton and Lee~Kristina Toutanova. 2019.
\newblock Bert: Pre-training of deep bidirectional transformers for language understanding.
\newblock In \emph{Proceedings of naacL-HLT}, volume~1, page~2.

\bibitem[{Lin et~al.(2021)Lin, Meng, Sun, Han, Kuang, Li, and Wu}]{lin-etal-2021-bertgcn}
Yuxiao Lin, Yuxian Meng, Xiaofei Sun, Qinghong Han, Kun Kuang, Jiwei Li, and Fei Wu. 2021.
\newblock \href {https://doi.org/10.18653/v1/2021.findings-acl.126} {{B}ert{GCN}: Transductive text classification by combining {GNN} and {BERT}}.
\newblock In \emph{Findings of the Association for Computational Linguistics: ACL-IJCNLP 2021}, pages 1456--1462, Online. Association for Computational Linguistics.

\bibitem[{Liu et~al.(2016)Liu, Qiu, and Huang}]{Liu2016RecurrentNN}
Pengfei Liu, Xipeng Qiu, and Xuanjing Huang. 2016.
\newblock \href {https://api.semanticscholar.org/CorpusID:16017905} {Recurrent neural network for text classification with multi-task learning}.
\newblock \emph{ArXiv}, abs/1605.05101.

\bibitem[{Liu et~al.(2019)Liu, Ott, Goyal, Du, Joshi, Chen, Levy, Lewis, Zettlemoyer, and Stoyanov}]{liu2019roberta}
Yinhan Liu, Myle Ott, Naman Goyal, Jingfei Du, Mandar Joshi, Danqi Chen, Omer Levy, Mike Lewis, Luke Zettlemoyer, and Veselin Stoyanov. 2019.
\newblock Roberta: A robustly optimized bert pretraining approach.
\newblock \emph{arXiv preprint arXiv:1907.11692}.

\bibitem[{Maas et~al.(2011)Maas, Daly, Pham, Huang, Ng, and Potts}]{maas2011learning}
Andrew Maas, Raymond~E Daly, Peter~T Pham, Dan Huang, Andrew~Y Ng, and Christopher Potts. 2011.
\newblock Learning word vectors for sentiment analysis.
\newblock In \emph{Proceedings of the 49th annual meeting of the association for computational linguistics: Human language technologies}, pages 142--150.

\bibitem[{Milano et~al.(2023)Milano, McGrane, and Leonelli}]{milano2023large}
Silvia Milano, Joshua~A McGrane, and Sabina Leonelli. 2023.
\newblock Large language models challenge the future of higher education.
\newblock \emph{Nature Machine Intelligence}, 5(4):333--334.

\bibitem[{Minaee et~al.(2021)Minaee, Kalchbrenner, Cambria, Nikzad, Chenaghlu, and Gao}]{minaee2021deep}
Shervin Minaee, Nal Kalchbrenner, Erik Cambria, Narjes Nikzad, Meysam Chenaghlu, and Jianfeng Gao. 2021.
\newblock Deep learning--based text classification: a comprehensive review.
\newblock \emph{ACM computing surveys (CSUR)}, 54(3):1--40.

\bibitem[{Moghimi et~al.(2016)Moghimi, Belongie, Saberian, Yang, Vasconcelos, and Li}]{moghimi2016boosted}
Mohammad Moghimi, Serge~J Belongie, Mohammad~J Saberian, Jian Yang, Nuno Vasconcelos, and Li-Jia Li. 2016.
\newblock Boosted convolutional neural networks.
\newblock In \emph{BMVC}, volume~5, page~6.

\bibitem[{OpenAI et~al.(2023)OpenAI, :, Achiam, Adler, and et~al.}]{openai2023gpt4}
OpenAI, :, Josh Achiam, Steven Adler, and Sandhini~Agarwal et~al. 2023.
\newblock \href {http://arxiv.org/abs/2303.08774} {Gpt-4 technical report}.

\bibitem[{Pang et~al.(2002)Pang, Lee, and Vaithyanathan}]{pang-etal-2002-thumbs}
Bo~Pang, Lillian Lee, and Shivakumar Vaithyanathan. 2002.
\newblock \href {https://doi.org/10.3115/1118693.1118704} {Thumbs up? sentiment classification using machine learning techniques}.
\newblock In \emph{Proceedings of the 2002 Conference on Empirical Methods in Natural Language Processing ({EMNLP} 2002)}, pages 79--86. Association for Computational Linguistics.

\bibitem[{Penedo et~al.(2023)Penedo, Malartic, Hesslow, Cojocaru, Cappelli, Alobeidli, Pannier, Almazrouei, and Launay}]{penedo2023refinedweb}
Guilherme Penedo, Quentin Malartic, Daniel Hesslow, Ruxandra Cojocaru, Alessandro Cappelli, Hamza Alobeidli, Baptiste Pannier, Ebtesam Almazrouei, and Julien Launay. 2023.
\newblock \href {http://arxiv.org/abs/2306.01116} {The refinedweb dataset for falcon llm: Outperforming curated corpora with web data, and web data only}.

\bibitem[{Qin et~al.(2023)Qin, Hu, Lin, Chen, Ding, Cui, Zeng, Huang, Xiao, Han et~al.}]{qin2023tool}
Yujia Qin, Shengding Hu, Yankai Lin, Weize Chen, Ning Ding, Ganqu Cui, Zheni Zeng, Yufei Huang, Chaojun Xiao, Chi Han, et~al. 2023.
\newblock Tool learning with foundation models.
\newblock \emph{arXiv preprint arXiv:2304.08354}.

\bibitem[{Raffel et~al.(2020)Raffel, Shazeer, Roberts, Lee, Narang, Matena, Zhou, Li, and Liu}]{raffel2020exploring}
Colin Raffel, Noam Shazeer, Adam Roberts, Katherine Lee, Sharan Narang, Michael Matena, Yanqi Zhou, Wei Li, and Peter~J Liu. 2020.
\newblock Exploring the limits of transfer learning with a unified text-to-text transformer.
\newblock \emph{The Journal of Machine Learning Research}, 21(1):5485--5551.

\bibitem[{Ren et~al.(2023)Ren, Zhou, Meng, Huang, Wang, Wang, Li, Zhang, Podolskiy, Arshinov et~al.}]{ren2023pangu}
Xiaozhe Ren, Pingyi Zhou, Xinfan Meng, Xinjing Huang, Yadao Wang, Weichao Wang, Pengfei Li, Xiaoda Zhang, Alexander Podolskiy, Grigory Arshinov, et~al. 2023.
\newblock Pangu-$\{$$\backslash$Sigma$\}$: Towards trillion parameter language model with sparse heterogeneous computing.
\newblock \emph{arXiv preprint arXiv:2303.10845}.

\bibitem[{Shen et~al.(2023)Shen, Hou, Zhou, Du, Longpre, Wei, Chung, Zoph, Fedus, Chen, Vu, Wu, Chen, Webson, Li, Zhao, Yu, Keutzer, Darrell, and Zhou}]{shen2023mixtureofexperts}
Sheng Shen, Le~Hou, Yanqi Zhou, Nan Du, Shayne Longpre, Jason Wei, Hyung~Won Chung, Barret Zoph, William Fedus, Xinyun Chen, Tu~Vu, Yuexin Wu, Wuyang Chen, Albert Webson, Yunxuan Li, Vincent Zhao, Hongkun Yu, Kurt Keutzer, Trevor Darrell, and Denny Zhou. 2023.
\newblock \href {http://arxiv.org/abs/2305.14705} {Mixture-of-experts meets instruction tuning:a winning combination for large language models}.

\bibitem[{Singhal et~al.(2023)Singhal, Azizi, Tu, Mahdavi, Wei, Chung, Scales, Tanwani, Cole-Lewis, Pfohl et~al.}]{singhal2023large}
Karan Singhal, Shekoofeh Azizi, Tao Tu, S~Sara Mahdavi, Jason Wei, Hyung~Won Chung, Nathan Scales, Ajay Tanwani, Heather Cole-Lewis, Stephen Pfohl, et~al. 2023.
\newblock Large language models encode clinical knowledge.
\newblock \emph{Nature}, 620(7972):172--180.

\bibitem[{Socher et~al.(2013)Socher, Perelygin, Wu, Chuang, Manning, Ng, and Potts}]{socher2013recursive}
Richard Socher, Alex Perelygin, Jean Wu, Jason Chuang, Christopher~D Manning, Andrew~Y Ng, and Christopher Potts. 2013.
\newblock Recursive deep models for semantic compositionality over a sentiment treebank.
\newblock In \emph{Proceedings of the 2013 conference on empirical methods in natural language processing}, pages 1631--1642.

\bibitem[{Sun et~al.(2022)Sun, He, Qian, Zhou, Huang, and Qiu}]{sun2022bbtv2}
Tianxiang Sun, Zhengfu He, Hong Qian, Yunhua Zhou, Xuan-Jing Huang, and Xipeng Qiu. 2022.
\newblock Bbtv2: towards a gradient-free future with large language models.
\newblock In \emph{Proceedings of the 2022 Conference on Empirical Methods in Natural Language Processing}, pages 3916--3930.

\bibitem[{Sun et~al.(2021)Sun, Wang, Feng, Ding, Pang, Shang, Liu, Chen, Zhao, Lu, Liu, Wu, Gong, Liang, Shang, Sun, Liu, Ouyang, Yu, Tian, Wu, and Wang}]{sun2021ernie}
Yu~Sun, Shuohuan Wang, Shikun Feng, Siyu Ding, Chao Pang, Junyuan Shang, Jiaxiang Liu, Xuyi Chen, Yanbin Zhao, Yuxiang Lu, Weixin Liu, Zhihua Wu, Weibao Gong, Jianzhong Liang, Zhizhou Shang, Peng Sun, Wei Liu, Xuan Ouyang, Dianhai Yu, Hao Tian, Hua Wu, and Haifeng Wang. 2021.
\newblock \href {http://arxiv.org/abs/2107.02137} {Ernie 3.0: Large-scale knowledge enhanced pre-training for language understanding and generation}.

\bibitem[{Touvron et~al.(2023)Touvron, Martin, Stone, Albert, Almahairi, Babaei, Bashlykov, Batra, Bhargava, Bhosale et~al.}]{touvron2023llama}
Hugo Touvron, Louis Martin, Kevin Stone, Peter Albert, Amjad Almahairi, Yasmine Babaei, Nikolay Bashlykov, Soumya Batra, Prajjwal Bhargava, Shruti Bhosale, et~al. 2023.
\newblock Llama 2: Open foundation and fine-tuned chat models.
\newblock \emph{arXiv preprint arXiv:2307.09288}.

\bibitem[{Wang and Banko(2021)}]{wang2021practical}
Cindy Wang and Michele Banko. 2021.
\newblock Practical transformer-based multilingual text classification.
\newblock In \emph{Proceedings of the 2021 Conference of the North American Chapter of the Association for Computational Linguistics: Human Language Technologies: Industry Papers}, pages 121--129.

\bibitem[{Wu et~al.(2024)Wu, Zhang, Han, Hou, Wang, Liu, Gong, and Ge}]{wu2024quartet}
Hui Wu, Yuanben Zhang, Zhonghe Han, Yingyan Hou, Lei Wang, Siye Liu, Qihang Gong, and Yunping Ge. 2024.
\newblock \href {http://arxiv.org/abs/2401.03158} {Quartet logic: A four-step reasoning (qlfr) framework for advancing short text classification}.

\bibitem[{Wu et~al.(2023)Wu, Irsoy, Lu, Dabravolski, Dredze, Gehrmann, Kambadur, Rosenberg, and Mann}]{wu2023bloomberggpt}
Shijie Wu, Ozan Irsoy, Steven Lu, Vadim Dabravolski, Mark Dredze, Sebastian Gehrmann, Prabhanjan Kambadur, David Rosenberg, and Gideon Mann. 2023.
\newblock Bloomberggpt: A large language model for finance.
\newblock \emph{arXiv preprint arXiv:2303.17564}.

\bibitem[{Wyner et~al.(2017)Wyner, Olson, Bleich, and Mease}]{wyner2017explaining}
Abraham~J Wyner, Matthew Olson, Justin Bleich, and David Mease. 2017.
\newblock Explaining the success of adaboost and random forests as interpolating classifiers.
\newblock \emph{The Journal of Machine Learning Research}, 18(1):1558--1590.

\bibitem[{Xiaofei et~al.(2023)Xiaofei, Xiaoya, Jiwei, Fei, and et~al.}]{pavlopoulos2023detecting}
Sun Xiaofei, Li~Xiaoya, Li~Jiwei, Wu~Fei, and et~al. 2023.
\newblock Text classification via large language models.
\newblock In \emph{Findings of the Association for Computational Linguistics: EMNLP 2023}, pages 8990--9005.

\bibitem[{Xie et~al.(2021)Xie, Huang, Du, Peng, and Nie}]{xie2021inductive}
Qianqian Xie, Jimin Huang, Pan Du, Min Peng, and Jian-Yun Nie. 2021.
\newblock Inductive topic variational graph auto-encoder for text classification.
\newblock In \emph{Proceedings of the 2021 Conference of the North American Chapter of the Association for Computational Linguistics: Human Language Technologies}, pages 4218--4227.

\bibitem[{Yang et~al.(2023)Yang, Liu, and Wang}]{yang2023fingpt}
Hongyang Yang, Xiao-Yang Liu, and Christina~Dan Wang. 2023.
\newblock \href {http://arxiv.org/abs/2306.06031} {Fingpt: Open-source financial large language models}.

\bibitem[{Yang et~al.(2019)Yang, Dai, Yang, Carbonell, Salakhutdinov, and Le}]{yang2019xlnet}
Zhilin Yang, Zihang Dai, Yiming Yang, Jaime Carbonell, Russ~R Salakhutdinov, and Quoc~V Le. 2019.
\newblock Xlnet: Generalized autoregressive pretraining for language understanding.
\newblock \emph{Advances in neural information processing systems}, 32.

\bibitem[{Yao et~al.(2018)Yao, Mao, and Luo}]{Yao2018GraphCN}
Liang Yao, Chengsheng Mao, and Yuan Luo. 2018.
\newblock \href {https://api.semanticscholar.org/CorpusID:52284222} {Graph convolutional networks for text classification}.
\newblock \emph{ArXiv}, abs/1809.05679.

\bibitem[{Yao et~al.(2019)Yao, Mao, and Luo}]{yao2019graph}
Liang Yao, Chengsheng Mao, and Yuan Luo. 2019.
\newblock Graph convolutional networks for text classification.
\newblock In \emph{Proceedings of the AAAI conference on artificial intelligence}, volume~33, pages 7370--7377.

\bibitem[{Ye et~al.(2023)Ye, Xu, Xu, Ye, Yan, Zhou, Wang, Hu, Shi, Shi, Li, Xu, Chen, Tian, Qi, Zhang, and Huang}]{ye2023mplugowl}
Qinghao Ye, Haiyang Xu, Guohai Xu, Jiabo Ye, Ming Yan, Yiyang Zhou, Junyang Wang, Anwen Hu, Pengcheng Shi, Yaya Shi, Chenliang Li, Yuanhong Xu, Hehong Chen, Junfeng Tian, Qian Qi, Ji~Zhang, and Fei Huang. 2023.
\newblock \href {http://arxiv.org/abs/2304.14178} {mplug-owl: Modularization empowers large language models with multimodality}.

\bibitem[{Zeng et~al.(2023)Zeng, Liu, Du, Wang, Lai, Ding, Yang, Xu, Zheng, Xia, Tam, Ma, Xue, Zhai, Chen, Liu, Zhang, Dong, and Tang}]{zeng2023glm-130b}
Aohan Zeng, Xiao Liu, Zhengxiao Du, Zihan Wang, Hanyu Lai, Ming Ding, Zhuoyi Yang, Yifan Xu, Wendi Zheng, Xiao Xia, Weng~Lam Tam, Zixuan Ma, Yufei Xue, Jidong Zhai, Wenguang Chen, Zhiyuan Liu, Peng Zhang, Yuxiao Dong, and Jie Tang. 2023.
\newblock \href {https://openreview.net/forum?id=-Aw0rrrPUF} {{GLM}-130b: An open bilingual pre-trained model}.
\newblock In \emph{The Eleventh International Conference on Learning Representations (ICLR)}.

\bibitem[{Zhang et~al.(2023{\natexlab{a}})Zhang, Chen, Jiang, Yu, Chen, Li, Chen, Wu, Zhang, Xiao et~al.}]{zhang2023huatuogpt}
Hongbo Zhang, Junying Chen, Feng Jiang, Fei Yu, Zhihong Chen, Jianquan Li, Guiming Chen, Xiangbo Wu, Zhiyi Zhang, Qingying Xiao, et~al. 2023{\natexlab{a}}.
\newblock Huatuogpt, towards taming language model to be a doctor.
\newblock \emph{arXiv preprint arXiv:2305.15075}.

\bibitem[{Zhang et~al.(2015)Zhang, Zhao, and LeCun}]{zhang2015character}
Xiang Zhang, Junbo Zhao, and Yann LeCun. 2015.
\newblock Character-level convolutional networks for text classification.
\newblock \emph{Advances in neural information processing systems}, 28.

\bibitem[{Zhang et~al.(2023{\natexlab{b}})Zhang, Ma, Tiwari, Zhang, Masud, Shorfuzzaman, and Song}]{zhang2023stance}
Yazhou Zhang, Dan Ma, Prayag Tiwari, Chen Zhang, Mehedi Masud, Mohammad Shorfuzzaman, and Dawei Song. 2023{\natexlab{b}}.
\newblock Stance-level sarcasm detection with bert and stance-centered graph attention networks.
\newblock \emph{ACM Transactions on Internet Technology}, 23(2):1--21.

\bibitem[{Zhang et~al.(2023{\natexlab{c}})Zhang, Wang, Wu, Tiwari, Li, Wang, and Qin}]{zhang2023dialoguellm}
Yazhou Zhang, Mengyao Wang, Youxi Wu, Prayag Tiwari, Qiuchi Li, Benyou Wang, and Jing Qin. 2023{\natexlab{c}}.
\newblock \href {http://arxiv.org/abs/2310.11374} {Dialoguellm: Context and emotion knowledge-tuned large language models for emotion recognition in conversations}.

\bibitem[{Zhao et~al.(2023)Zhao, Zhou, Li, Tang, Wang, Hou, Min, Zhang, Zhang, Dong, Du, Yang, Chen, Chen, Jiang, Ren, Li, Tang, Liu, Liu, Nie, and Wen}]{zhao2023survey}
Wayne~Xin Zhao, Kun Zhou, Junyi Li, Tianyi Tang, Xiaolei Wang, Yupeng Hou, Yingqian Min, Beichen Zhang, Junjie Zhang, Zican Dong, Yifan Du, Chen Yang, Yushuo Chen, Zhipeng Chen, Jinhao Jiang, Ruiyang Ren, Yifan Li, Xinyu Tang, Zikang Liu, Peiyu Liu, Jian-Yun Nie, and Ji-Rong Wen. 2023.
\newblock \href {http://arxiv.org/abs/2303.18223} {A survey of large language models}.

\bibitem[{Zhou et~al.(2016)Zhou, Wan, and Xiao}]{zhou2016attention}
Xinjie Zhou, Xiaojun Wan, and Jianguo Xiao. 2016.
\newblock Attention-based lstm network for cross-lingual sentiment classification.
\newblock In \emph{Proceedings of the 2016 conference on empirical methods in natural language processing}, pages 247--256.

\end{thebibliography}

\bibliographystyle{acl_natbib}

\section{Explanations of The Complementary and Robustness Across Base Learners}\label{sec:appa}

The complementarity among multiple base learners, as observed in ensemble learning frameworks like boosting, refers to the ability of different foundational models to recognize and process distinct features or patterns within the data. For RGPT, which employs LLMs as base learners, this complementarity is manifested in several aspects:

\textbf{(1) Feature space coverage.} Each fine-tuned LLM may exhibit varying degrees of understanding and capturing capabilities for different semantic, syntactic structures, or contextual information in the input text. For instance, one LLM may excel at handling long-distance dependency relationships, while another may demonstrate greater accuracy in understanding domain-specific terms.

\textbf{(2) Error distribution.} As the sample weights are adjusted based on the prediction errors of the preceding weak learners during each iteration, subsequent learners focus more on the previously misclassified samples. Consequently, even if the foundational architectures of all LLMs are similar, they address and correct different subsets of data, creating complementarity.

\textbf{(3) Randomness and robustness.} Despite fine-tuning for the same task, different initialization states and random factors during the training process (such as the stochastic nature of gradient descent) may lead LLMs to produce distinct decision boundaries. These boundaries may intersect or misalign in complex data distributions, enhancing the overall robustness and generalization performance of the ensemble model.

\textbf{(4) Model capacity.} While LLMs possess high capacity, a single model may not fully leverage all its parameters to adapt to complex tasks, especially with limited training data. Through multiple rounds of fine-tuning and ensemble combination, the model potential can be better explored, allowing each learner to focus on specific aspects of the task, resulting in overall optimization.

\section{Recurrent Ensembling The Base Learners: Algorithm and Illustration }
Here, we present further details of RGPT in Algorithm~\ref{alg:rgpt} and the overall architecture of ensembling in Fig.~\ref{fig:prompt}
\begin{algorithm}[htp]
    \caption{Recurrent ensemble Learning of RGPT}
    \label{alg:rgpt}
\small
    \begin{algorithmic}[1]
        \Require
            \State \textbf{Input:}
            \State \hspace{\algorithmicindent} $\mathcal{D}^{(0)}$: Original training dataset with $N$ samples $(x_i^{(0)}, y_i^{(0)})$
            \State \hspace{\algorithmicindent} $\mathcal{LM}_0$: LLaMA~2 as initial base learner
            \State \hspace{\algorithmicindent} $K$: Number of base learners

        \Ensure
            \State \textbf{Output:}
            \State \hspace{\algorithmicindent} $\mathcal{M}_{ensemble}$: Recursively ensembled model

         \State \textbf{Training:}
            \State  \hspace{\algorithmicindent} Initialize data weights $\mathcal{W}^{(0)} = \left \{ w_1^{(0)}, \ldots, w_N^{(0)} \right \}$ where $w_i^{(0)} = \frac{1}{N}, \forall i \in N$

        \For{$k = 1, 2, \ldots, K$}
            \State Construct prompt $\text{Prompt}_i^{(k)}=  {INS}_i \oplus x_i^{\left ( k \right ) }$
            \State Fine-tune $\mathcal{LM}_k$ with weighted training samples:
                \[
                    \mathcal{LM}_k = argmin_{\theta^{(k)}} \sum_{\mathcal{D}^{(k)}} w_i^{(k)} \cdot \mathcal{L}(y_i^{(k)}, \text{f}_k(x_i^{(k)}; \theta^{(k)}))
                \]
            \State Compute error rate $\epsilon^{(k)}$ of $\mathcal{LM}_k$
            
            \State Calculate weight coefficient $\alpha^{(k)} = \log\frac{1-\epsilon^{(k)}}{\epsilon^{(k)}} + \log(c-1)$
            \State Update data weights for ${k+1}^{th}$ iteration:
                \[
                    \mathcal{W}^{(k+1)}_i =
                        \begin{cases}
                            \frac{w_i^{(k)}}{Z_k} e^{-\alpha^{(k)}} & \text{if } \mathcal{LM}_k(x_i^{(k)}) = y_i^{(k)} \\
                            \frac{w_i^{(k)}}{Z_k} e^{\alpha^{(k)}} & \text{if } \mathcal{LM}_k(x_i^{(k)}) \neq y_i^{(k)}
                        \end{cases}
                \]
                \State Normalize weights by $Z_k$ to ensure $\sum_{i=1}^N w_i^{(k+1)} = 1$
                \EndFor
                
                   \State \textbf{Inference:}
                   \For{$k = 1, 2, \ldots, K$}
\State Forward the prompt through $k^{th}$ base learner $\mathcal{LM}_k$
\State Obtain the classification result $\hat{y}^{(k)}_i$
            \State Update prompt for next iteration: 
                \[
                    \text{Prompt}_i^{(k+1)} = \text{Prompt}_i^{(k)} \oplus \{\hat{y}^{(k)}_i, \epsilon^{(k)}\}
                \]
                
        \EndFor
        
        \State \Return $\mathcal{M}_{ensemble}=F(\mathcal{LM}_1, \mathcal{LM}_2, \ldots, \mathcal{LM}_K)$
    \end{algorithmic}
\end{algorithm}

\end{document}